\documentclass[final,3p,times,twocolumn]{elsarticle}

\usepackage{amssymb}
\usepackage{amssymb}
\usepackage{amsthm}
\usepackage{algorithm}
\usepackage{multirow}
\usepackage{algorithmic}
\usepackage{amsmath}
\usepackage{array}
\usepackage{color}
\usepackage{multirow}

\begin{document}

\begin{frontmatter}
\title{Reliable Object Tracking by Multimodal Hybrid Feature Extraction and Transformer-Based Fusion}

\author[label1]{Hongze Sun}
\author[label1]{Rui Liu}
\author[label1]{Wuque Cai}
\author[label1]{Jun Wang}
\author[label1]{Yue Wang}
\author[label3]{Huajin Tang}
\author[label1,label5]{Yan Cui}
\author[label1,label6]{Dezhong Yao\corref{cor1}}
\author[label1]{and Daqing Guo\corref{cor1}}
\address[label1]{Clinical Hospital of Chengdu Brain Science Institute, MOE Key Lab for NeuroInformation, School of Life Science and Technology, University of Electronic Science and Technology of China, Chengdu 611731, China.}
\address[label3]{College of Computer Science and Technology, Zhejiang University, Hangzhou 310027, China.}
\address[label5]{Sichuan Academy of Medical Sciences and Sichuan Provincial People's Hospital, Chengdu 611731, China.}
\address[label6]{Research Unit of NeuroInformation (2019RU035), Chinese Academy of Medical Sciences, Chengdu 611731, China.}
\cortext[cor1]{Corresponding authors: dyao@uestc.edu.cn (Dezhong Yao) and dqguo@uestc.edu.cn (Daqing Guo).}

\begin{abstract}
Visual object tracking, which is primarily based on visible light image sequences, encounters numerous challenges in complicated scenarios, such as low light conditions, high dynamic ranges, and background clutter. To address these challenges, incorporating the advantages of multiple visual modalities is a promising solution for achieving reliable object tracking. However, the existing approaches usually integrate multimodal inputs through adaptive local feature interactions, which cannot leverage the full potential of visual cues, thus resulting in insufficient feature modeling. In this study, we propose a novel multimodal hybrid tracker~(MMHT) that utilizes frame-event-based data for reliable single object tracking. The MMHT model employs a hybrid backbone consisting of an artificial neural network~(ANN) and a spiking neural network~(SNN) to extract dominant features from different visual modalities and then uses a unified encoder to align the features across different domains. Moreover, we propose an enhanced transformer-based module to fuse multimodal features using attention mechanisms. With these methods, the MMHT model can effectively construct a multiscale and multidimensional visual feature space and achieve discriminative feature modeling. Extensive experiments demonstrate that the MMHT model exhibits competitive performance in comparison with that of other state-of-the-art methods. Overall, our results highlight the effectiveness of the MMHT model in terms of addressing the challenges faced in visual object tracking tasks.
\end{abstract}

\begin{keyword}
Object tracking \sep Multimodal fusion \sep Spiking neural networks \sep Transformer.
\end{keyword}
\end{frontmatter}

\section{Introduction}
Visual object tracking is a fundamental yet challenging computer vision task that  has a wide range of applications in the real world~\citep{liu2021overview,luo2021multiple}. Benefiting from the progress achieved with respect to deep neural networks and big data, trackers based on feature modeling and end-to-end training have become the mainstream models for solving single object tracking problems~\citep{danelljan2019atom, bhat2019learning}. To address challenging visual scenes, researchers have recently proposed introducing more task-oriented visual cues by  integrating multimodal information (such as thermal, depth and event information), thereby enhancing the robustness of feature modeling~\citep{yang2023resource, zhu2023visual, hui2023bridging, zhang2023efficient}. Among these visual information modalities, event data recorded by bioinspired event cameras are attracting increasing attention~\citep{zhang2023frame, messikommer2023data, zhang2022spiking}.

\begin{figure}[t]
\centering
\includegraphics[width=1.0\columnwidth]{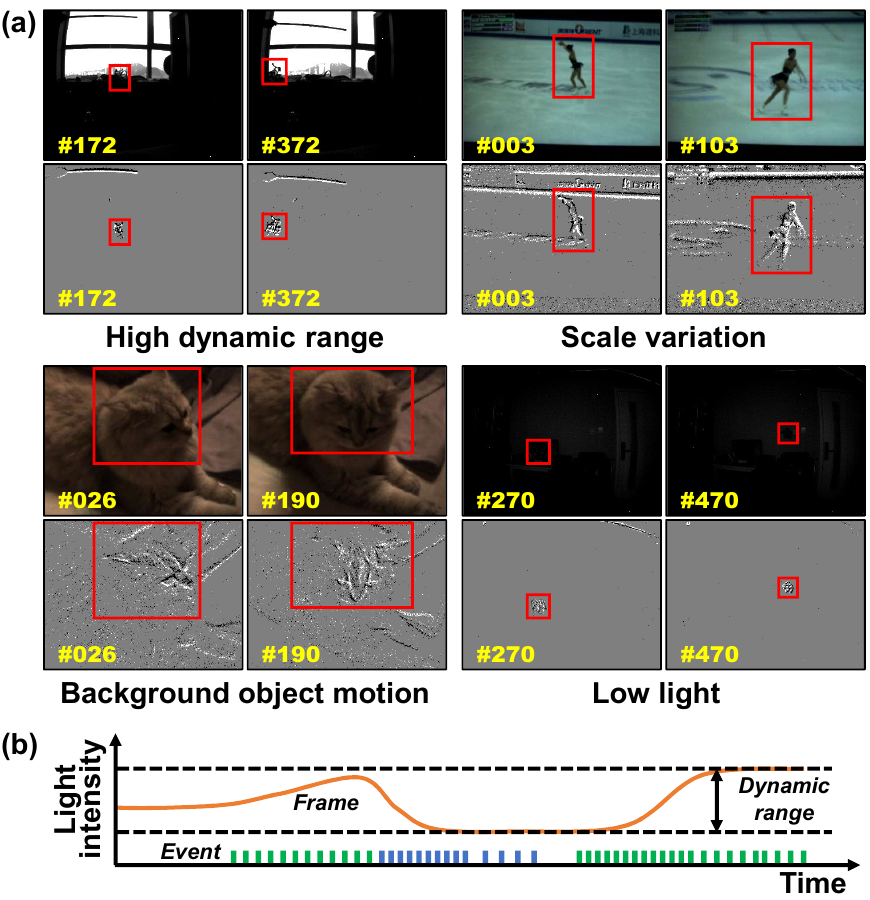} 
\caption{(a) Complementary characteristics of frame- and event-based images. Event-based cameras excel in challenging conditions, such as environments with high dynamic ranges and low light, while frame-based cameras enable the capture of rich detailed information.
	(b) Schematic of the imaging principle. Frame-based cameras synchronously record light intensity, while event-based cameras utilize ON/OFF spike trains to asynchronously reflect light intensity changes. Additionally, event-based cameras are compatible with  higher dynamic ranges than those of frame-based cameras.}
\label{Figure1}
\end{figure}

In contrast with conventional frame-based cameras that employ light intensity to construct spatial appearance information, event cameras record light intensity changes in the temporal domain~(as shown in Fig.~\ref{Figure1}), enabling the representation of sparse spatiotemporal information~\citep{9869656}. Due to their higher dynamic ranges and sampling frequencies, event cameras demonstrate inherent proficiency in challenging conditions, including environments with low light, scenes with high dynamic ranges, spatiotemporal coupling scenarios, and active filtering tasks~\citep{li2019event}. In recent years, numerous event-based datasets have been published and utilized for various visual tasks, such as classification, recognition, optical flow estimation, object tracking and semantic segmentation~\citep{10032591, Cai2022ASA,zhu2018multivehicle,li2017cifar10, ji2023sctn}. In particular, large-scale frame-event-based datasets constructed using event cameras offer opportunities to perform fusion studies in related fields~\citep{yang2023event}. In this study, we leverage the strengths of both the event and frame modalities to achieve reliable object tracking.

To fully leverage the potential of multimodal data, two key challenges should be effectively addressed. (1) It is crucial to devise a hybrid feature extraction network that enables the targeted exploration of the visual cues within frame-event-based inputs. (2) To facilitate discriminative feature modeling, a novel framework for performing feature alignment and fusion is essential. We note that several studies have begun to explore the relevant problems. To process data with diverse modalities, researchers have proposed specific task-oriented networks for feature extraction purposes~\citep{zhang2021object, zhang2023efficient}. However, these networks generally  possess complex architectures and information interactions. Furthermore, a simple yet efficient approach for decoupling the spatiotemporal features contained in event data is still lacking, which also imposes a bottleneck on the subsequent feature fusion process. Researchers must integrate all visual cues through local communication conducted on separate information channels rather than in a unified feature space. This may result in fusion modules that lack global awareness and are unable to fairly assess the relationships between different features.

To address the above challenges, we propose a novel multimodal hybrid tracker~(MMHT) to effectively integrate information from two visual modalities for reliable object tracking. We develop two key components in the MMHT: multimodal feature extraction~(MMFE) and transformer-based feature fusion~(TFF) modules. (1) MMFE: Conventional artificial neural networks~(ANNs) have demonstrated robust and excellent capabilities in terms of processing the visible light modality~\citep{NIPS2015_14bfa6bb, ye2022joint, song2021efficient}. Brain-inspired spiking neural networks~(SNNs) have the ability to synchronously perceive spatiotemporal features, making them suitable for neuromorphic event data~\citep{yu2023brain, pei2019towards}. Therefore, we propose a hybrid network that combines ANNs and SNNs to extract multimodal features, thereby constructing a multiscale and multidimensional visual representation space. By introducing a newly developed synapse-threshold synergistic learning approach for SNNs~\citep{sun2023synapse}, the MMFE module can optimize the network parameters in an end-to-end manner and achieve excellent performance. (2) TFF: Recently, the transformer architecture, which utilizes self-attention for global information modeling, has demonstrated remarkable capabilities in various intelligent tasks and has gained increasing attention~\citep{vaswani2017attention, dosovitskiy2020image, brown2020language}. Importantly, transformers are applicable to different modalities, providing a concise and efficient unified framework for multimodal fusion. Here, we construct the TFF module based on enhanced transformers by introducing a cross-attention mechanism. With the TFF module, we can align the visual representations derived from different modalities and achieve feature modeling across different domains. Accordingly, the superiority of the MMFE and TFF modules establishes a foundation for achieving reliable object tracking.

Our contributions in this paper are presented as follows:
\begin{itemize}
\item We propose a novel MMHT model for reliably performing single object tracking by jointly exploiting the frame and event domains.
\item We design a hybrid ANN-SNN frame-event-based feature extraction approach to construct a multiscale and multidimensional visual representation space.
\item We develop an enhanced transformer-based feature fusion strategy that operates across domains to perform discriminative feature modeling.
\item Experiments show that the MMHT model achieves competitive performance in
comparison with that of other state-of-the-art models on challenging benchmark datasets (FE108, COESOT and VisEvent).
\end{itemize}

\section{Related Works}
In this section, we briefly review the recent works conducted on multimodal object tracking, multimodal feature modeling and frame-event-based object tracking, which are highly associated with our study.

\subsection{Multimodal Object Tracking}
To cope with complex scenarios, an increasing number of modalities are being incorporated into object tracking tasks to enable robust and comprehensive feature modeling. Currently, the most valuable modalities for research include the thermal, depth, event, and language modalities~\citep{yu2023brain, zhu2023visual, li2022dual}. The thermal modality detects the surface temperature distribution of an object through thermal radiation, and its imaging process remains unaffected by weather conditions. Therefore, the thermal modality is often employed to complete tracking tasks in extreme weather conditions. However, the thermal modality also has drawbacks in terms of resolution and noise, thereby making RGB-T fusion a popular research topic in the object tracking field~\citep{xiao2022attribute}. The depth modality constructs a 3D spatial relationship by recording the distances from objects to the camera, exhibiting excellent representation capabilities for cases with object occlusion. RGB-D fusion has demonstrated a significant impact in fields such as autonomous driving and facial detection~\citep{yang2023resource}. In contrast, an event camera captures light intensity changes at a high frequency and exhibits high sensitivity to object motion. By incorporating the event modality, trackers can obtain stable object-oriented spatiotemporal features~\citep{9869656}. Therefore, researchers have begun exploring frame-event-based tracking, and several large-scale datasets have been released to validate the performance of the developed trackers~\citep{wang2021visevent, tang2022revisiting, zhang2021object}.

\subsection{Multimodal Feature Modeling}
To achieve reliable feature modeling through complementary advantages, a multimodal model typically consists of three components: feature extraction, feature alignment, and feature fusion modules. The most common feature extraction strategy involves using specialized dual-stream architectures inspired by prior knowledge based on different modalities~\citep{yu2023brain, zhang2021object,jiang2023cmci}.  This approach circumvents the challenge of designing models for inconsistent data formats while ensuring the completeness and relevance of the feature extraction process. Although some approaches consciously align their features during the extraction stage, most models still require a feature space transformation as a foundation for performing feature fusion. Generally, adaptive feature modulation and high-dimensional kernel projection are common techniques employed for feature alignment~\citep{zhang2023frame, yu2023brain}. In terms of feature fusion, a variety of techniques, ranging from simple feature combinations or concatenations to attention-based enhancements, have proven to be effective at leveraging the advantages of multimodal data~\citep{wang2021visevent}. Furthermore, inspired by transformer-based architectures, some researchers have begun exploring the possibility of constructing a unified framework~\citep{zhu2023visual, tang2022revisiting}. Their aim is to utilize an improved transformer, that encompasses all the aforementioned steps to directly accomplish feature modeling. 

\subsection{Frame-Event-Based Object Tracking}
Tracking methods based on frame-event-based modalities have demonstrated remarkable capabilities in terms of leveraging extreme lighting conditions and extracting detailed texture information~\citep{zhang2023frame, yang2019dashnet, zhang2021object, zhu2023visual, el2022high}. However, due to the structural differences between the information representations of these two modalities, it is crucial to design architectures that can effectively explore potential complementary task-oriented features. 

\begin{figure*}[t]
\centering
\includegraphics[width=1.0\textwidth]{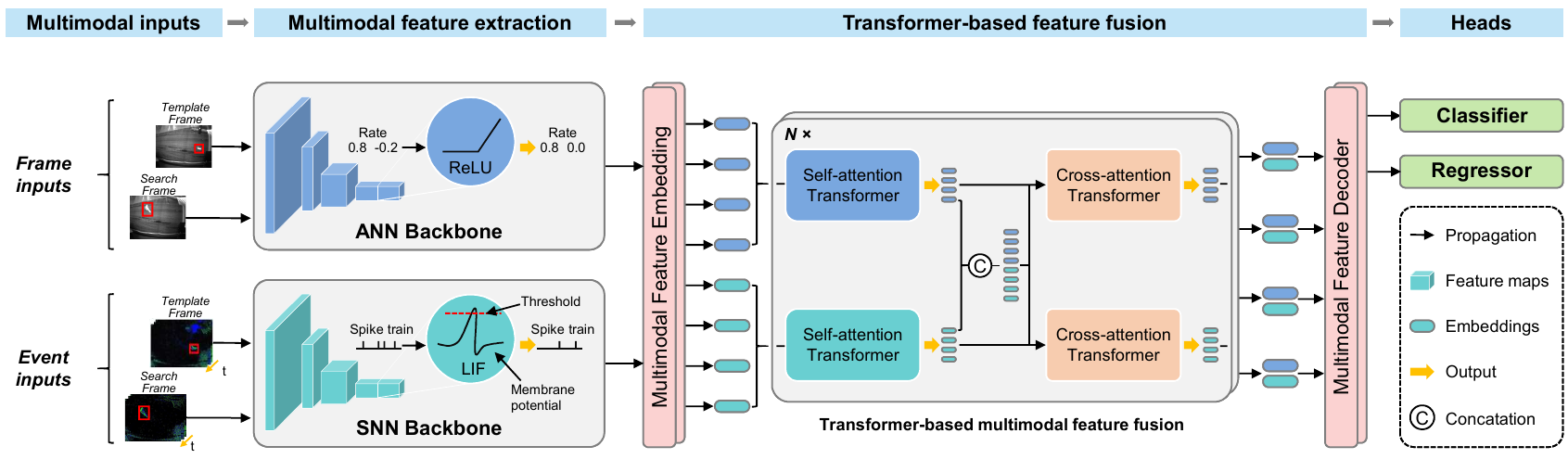} 
\caption{The overall framework of the proposed MMHT. Frame- and event-modality inputs are initially processed by hybrid backbones to extract discriminative features. These features are subsequently embedded as patch embeddings using the multimodal feature embedding module, enabling effective cross-modal visual cue alignment. The proposed transformer-based multimodal feature fusion blocks leverage diverse attention modules to enhance and seamlessly integrate cross-domain features. Ultimately, the multimodal feature decoder produces fusion-level inputs, which are employed by our heads to perform accurate object tracking.}
\label{Figure2}
\end{figure*}

Inspired by the hypothetical two-stream model of visual neural processing~\citep{goodale1992separate}, prior studies have proposed various two-branch architectures for performing targeted feature extraction in different modalities. To achieve effective environmental perception in extreme scenarios, some existing approaches consider the event-based modality as a complement to the conventional frame-based modality. These methods focus on implementing cross-modal enhancements at the feature level~\citep{zhang2023frame, wang2021visevent, wang2023sstformer} or employ hybrid architectures that combine SNNs, ANNs and hard attention mechanisms to facilitate efficient feature interaction~\citep{yang2019dashnet, zhao2022framework, wang2023sstformer}. Other works have aimed to provide more reliable feature cues for object tracking by delving into the temporal properties of the event modality using recurrent neural network (RNN)-like structures~\citep{zhang2021object}. This strategy extends the scope of feature mining from the original spatial dimension to the spatiotemporal dimension. 

As with other studies concerning multimodal object tracking, some researchers have highlighted the simultaneous extraction of frame-event-based features using a unified framework structured with transformers~\citep{tang2022revisiting, zhu2023visual, zeng2023swineft}. Remarkably, these models have exhibited competitive performance on several mainstream large-scale datasets in comparison with that of two-branch models. For instance, a novel plug-and-play mask modeling strategy has been developed in a recent study~\citep{zhu2023cross}. By combining with a pretrained vision transformer, this strategy can led to notable performance enhancements for unified frameworks in object tracking tasks.

\section{Methods}
\subsection{Overview}
We first provide an overview of the proposed MMHT model. Briefly, the MMHT model evolves on the basis of discriminative correlation filter trackers, which are characterized by a shared target-specific feature extraction network and available online learning  heads~\citep{danelljan2019atom, bhat2019learning}. To achieve superior multimodal feature modeling capabilities, we propose a pioneering hybrid architecture that can effectively capture cross-domain visual cues. As depicted in Fig.~\ref{Figure2}, the framework of the MMHT model includes four parts: multimodal inputs, a multimodal feature extraction module, a transformer-based feature fusion module, and heads. Notably, the MMHT model is trainable in an end-to-end trainable manner. 

\subsection{Multimodal Inputs}
The bioinspired neuromorphic camera facilitates the simultaneous acquisition of frame-event-based data. The conventional frame-based input $f_{i}(x,y)$ captures the light intensity at the $i$-th exposure time $T_{i}$, where $(x,y)$ denotes the pixel location. In contrast, the event-based inputs $\left \{ [x_{k},y_{k},t_{k},p_{k}]  \right \}_{k=1}^{K}$ asynchronously record light intensity changes with their polarity ($ p_{k} \in \left \{-1, +1 \right \} $). $K$ denotes the number of events, and $t_{k}$ is the corresponding timestamp of the $k$-th event. For convenience, the event inputs are aggregated into a frame-based representation $g_{i,j}(x,y) $ as depicted in the following formulas:
\begin{equation}
\begin{array}{l}
	g_{i,j}(x,y) = [p_{k} \times \delta (t_{\rm max}-t_{k})+1] \times 127, \\
\end{array}
\label{eq1}
\end{equation}
with 
\begin{equation}
\begin{array}{l}
	t_{\rm max} = {\rm max}\left \{-1, t_{k} \times \delta (x-x_{k}, y-y_{k}) \right \},\\
\end{array}
\label{eq1a}
\end{equation}
subjected to $\forall t_{k} \in \left [T_{i}+jB, T_{i}+(j+1)B\right ]$. In Eqs.~(\ref{eq1}) and~(\ref{eq1a}), $g_{i,j}(x,y)$ represents the $j$-th aggregated frame during time period $[T_{i}, T_{i+1}]$~\citep{zhang2021object}, $B=\left ( T_{i+1}-T_{i} \right )/N $ denotes a time window with a temporal resolution of $N$, and $\delta$ is the Dirac delta function. In our present study, multimodal inputs are represented as a series of combinations $[(f_{i}, g_{i,1},\dots ,g_{i,N})]_{i=0}^{I}$.

\subsection{Multimodal Feature Extraction}
It is widely acknowledged that frame-based images possess the ability to objectively capture abundant texture information, thereby offering valuable visual cues for spatial feature modeling~\citep{deng2009imagenet}. Nevertheless, event-based data capture object-oriented edge and motion information across the spatiotemporal domain~\citep{yang2023event}, and both types of information are of equal importance for object tracking tasks. To efficiently extract diverse visual cues from multimodal inputs, we propose a novel hybrid backbone constructed with convolutional neural networks based on distinct types of neurons.

\subsubsection{ANN Backbones for the Frame Modality}
The pretrained ResNet18~(structured with conv1, conv2\_x, conv3\_x, conv4\_x, conv5\_x and fully connected layers) model demonstrates powerful transferability in downstream visual tasks~\citep{he2016deep}. Thus, we adopt the convolutional layers of ResNet18 as backbones for the frame modality. The feature maps generated by conv3\_x and conv4\_x are used as the low-level and high-level features~($F_{\rm l}^{i}$, $F_{\rm h}^{i}$), respectively.
\begin{equation}
F_{\rm l}^{i} = {\rm conv3\_x}({\rm conv2\_x}({\rm conv1}(f_{i}))),
\label{eq2}
\end{equation}
\begin{equation}
F_{\rm h}^{i} = {\rm conv4\_x}(F_{\rm l}^{i}).
\label{eq3}
\end{equation}

\begin{algorithm}[t]
\caption{Multimodal Feature Embedding}
\label{algorithm1}
\textbf{Inputs}: Feature maps $X^{C \times H \times W}$   \\
\textbf{Parameters}: Feature patch's resolution $p$ and embedding dimensionality $D_{{\rm dim}}$ \\
\textbf{Output}: The embedding
\begin{algorithmic}[1]
	\STATE Reshape $X^{C \times H \times W}$ to $X^{N \times (p^{2}\cdot C)}$  \ \ \ \ \ \ \ \ \ {\rm \#} $N=H\cdot W/p^{2}$\\
	\STATE $X^{N \times (p^{2}\cdot C)} \gets {\rm LayerNorm}(p^{2}\cdot C):X^{N \times (p^{2}\cdot C)}$ \\
	\STATE $X^{N \times D_{{\rm dim}}} \gets {\rm Linear}(p^{2}\cdot C, D_{{\rm dim}}):X^{N \times (p^{2}\cdot C)}$\\
	\STATE $X^{N \times D_{{\rm dim}}} \gets {\rm LayerNorm}(D_{dim}):X^{N \times D_{{\rm dim}}}$\\
	\STATE $X^{N \times D_{{\rm dim}}} \gets {\rm Dropout}:X^{N \times D_{{\rm dim}}}$
	\STATE \textbf{return} $X_{{\rm embed}}^{N \times D_{{\rm dim}}}$
\end{algorithmic}
\end{algorithm}

\begin{algorithm}[t]
\caption{Transformer-Based Multimodal Feature Fusion}
\label{algorithm2}
\textbf{Inputs}: Embeddings $F_{{\rm embed}}$ and $G_{{\rm embed}}$ \\
\textbf{Parameters}: Self-attention transformer blocks $\rm sat_{1}$, $\rm sat_{2}$, cross-attention transformer blocks $\rm cat_{1}$, $\rm cat_{2}$, and the number $I$ of TMFF modules \\
\textbf{Outputs}: Fusion embedding $T_{{\rm embed}}$
\begin{algorithmic}[1]
	\STATE \textbf{For} $i = 1$ to $I$ do:
	\STATE \ \ \ $F_{{\rm embed}} \gets {\rm sat}_{1}(F_{{\rm embed}})$\\
	\STATE \ \ \ $G_{{\rm embed}} \gets {\rm sat}_{2}(G_{{\rm embed}})$\\
	\STATE \ \ \ $D_{{\rm embed}} = {\rm concat}(F_{{\rm embed}},G_{{\rm embed}})$ \\
	\STATE \ \ \ $F_{{\rm embed}} \gets {\rm cat}_{1}(F_{{\rm embed}},D_{{\rm embed}})$\\
	\STATE \ \ \ $G_{{\rm embed}} \gets {\rm cat}_{2}(G_{{\rm embed}},D_{{\rm embed}})$\\
	\STATE \textbf{end for}
	\STATE $T_{{\rm embed}} = {\rm concat}(F_{{\rm embed}},G_{{\rm embed}})$ \\
	\STATE \textbf{return} $T_{{\rm embed}}$
\end{algorithmic}
\end{algorithm}

\begin{algorithm}[t]
\caption{Multimodal Feature Decoder}
\label{algorithm3}
\textbf{Inputs}: Embeddings $X^{2N \times D_{{\rm dim}}}$   \\
\textbf{Parameters}: Feature map resolution $W$ and $H$ \\
\textbf{Output}: Feature maps
\begin{algorithmic}[1]
	\STATE $X^{2N \times D_{{\rm dim}}} \gets {\rm LayerNorm}(D_{{\rm dim}}):X^{2N \times D_{{\rm dim}}}$ \\
	\STATE Reshape $X^{2N \times D_{{\rm dim}}}$ to $X^{D_{{\rm dim}} \times 2N}$  \\
	\STATE $X^{D_{{\rm dim}} \times H\cdot W} \gets {\rm Linear}(2N, HW):X^{D_{{\rm dim}} \times 2N}$\\
	\STATE $X^{D_{{\rm dim}} \times H\cdot W} \gets {\rm LayerNorm}(HW):X^{D_{{\rm dim}} \times H\cdot W}$ \\
	\STATE $X^{D_{{\rm dim}} \times H\cdot W} \gets {\rm Dropout}:X^{D_{{\rm dim}} \times H\cdot W}$ \\
	\STATE Reshape $X^{D_{{\rm dim}} \times HW}$ to $X^{D_{{\rm dim}} \times H \times W}$  \\
	\STATE \textbf{return} $X^{D_{{\rm dim}} \times H \times W}$
\end{algorithmic}
\end{algorithm}

\subsubsection{SNN Backbones for the Event Modality}
SNNs form a new generation of neural network models that leverage bioinspired neurons and discrete spike trains to mimic the intricate spatiotemporal dynamic processes observed in the human brain~\citep{maass1997networks}. SNNs have garnered significant attention due to their exceptional proficiency in extracting spatiotemporal features~\citep{10032591, zhang2022spiking, ma2022deep}. In this study, we employ the leaky integrate and fire~(LIF) neuron, a computational model that strikes a balance between dynamic complexity and computational simplicity, to construct SNN backbones for extracting features from the event modality. Without loss of generality, the iterative form of the LIF neuron utilized in our work can be described as follows:
\begin{equation}
u^{t} = \alpha\cdot (1-o^{t-1}) \cdot u^{t-1} + \textstyle \sum_{m=1}^{M}w_{m}\cdot o_{m}^{t},
\label{eq4}
\end{equation}
\begin{equation}
o^{t} = \sigma (u^{t}-u_{\rm th}),
\label{eq5}
\end{equation}
where $u^{t}$ and $o^{t}$ represent the neuronal membrane potential and spike output at time $t$, respectively. In addition, $\alpha$ and $u_{\rm th}$ are intrinsic neuronal properties: the membrane decay constant and spike threshold, respectively. $w_{m}$ denotes the synaptic weight. In this study, we use a novel proposed synapse-threshold synergistic learning approach to simultaneously train $w_{m}$ and $u_{\rm th}$ for SNNs~\citep{sun2023synapse}.

The architectures of the backbones comprise convolutional layers~(convl\_x, convh\_x) structured in the form of the feature extraction component in AlexNet~\citep{krizhevsky2012imagenet}.  These architectures undergo meticulous refinement and optimization to suit a variety of datasets and accommodate feature fusion modules~(detailed parameters are listed in Tab.\ref{table0}). To obtain low-level and high-level spiking feature trains~($[G_{{\rm l},1}^{i} ,\dots ,G_{{\rm l},N}^{i}]$, $[G_{{\rm h},1}^{i} ,\dots ,G_{{\rm h},N}^{i}]$), the following procedure is employed:
\begin{equation}
[G_{{\rm l},1}^{i} ,\dots ,G_{{\rm l},N}^{i}] = {\rm convl\_x}([g_{i,1},\dots g_{i,N}]),
\label{eq6}
\end{equation}
\begin{equation}
[G_{{\rm h},1}^{i} ,\dots ,G_{{\rm h},N}^{i}] = {\rm convh\_x}([G_{{\rm l},1}^{i} ,\dots ,G_{{\rm l},N}^{i}]).
\label{eq7}
\end{equation}
Utilizing the average firing rate observed over $\left [T_{i},T_{i+1} \right ]$, we code and normalize the spiking feature maps as feature maps $G_{\rm l}^{i}$ and $G_{\rm h}^{i}$:
\begin{equation}
G_{\rm l}^{i} = \frac{1}{N}{\textstyle \sum_{n=1}^{N}} G_{{\rm l},n}^{i},
\label{eq8}
\end{equation}
\begin{equation}
G_{\rm h}^{i} = \frac{1}{N}{\textstyle \sum_{n=1}^{N}} G_{{\rm h},n}^{i}.
\label{eq9}
\end{equation}
To address the nondifferential nature of spiking events, we employ an approximate gradient function during the feedback propagation process~\citep{wu2018spatio, wu2019direct}:
\begin{equation}
\sigma ^{\prime } = {\rm ReLU} \left (1-\left | x \right |   \right ),
\label{eq10}
\end{equation}
where $\rm ReLU$ represents the activation function of the rectified linear unit.

\subsection{Transformer-Based Feature Fusion}
Our proposed method aims to efficiently fuse visual cues in the complete feature space through an improved transformer-based module.

\subsubsection{Multimodal Feature Embedding}
To obtain modality-independent formalized embeddings, we introduce a novel feature embedding process in our approach. It consists mainly of reshaping conversion operations and a linear layer, which can effectively convert the original features into constant latent vectors while mitigating any potential inductive bias~\citep{vaswani2017attention}. The specific process and parameters are outlined in Algorithm~\ref{algorithm1} and Tab.\ref{table0}, respectively. Therefore, the feature maps derived from different modalities~($G_{\rm l}^{i}$, $F_{\rm l}^{i}$, $G_{\rm h}^{i}$ and $F_{\rm h}^{i}$) are transformed into uniform embeddings~($G_{\rm l,embed}^{i}$, $F_{\rm l,embed}^{i}$, $G_{\rm h,embed}^{i}$ and $F_{\rm h,embed}^{i}$). Note that the numbers and dimensions of the embeddings are equivalent across different modalities within our study.

\subsubsection{Transformer-Based Multimodal Feature Fusion}
The framework of the transformer-based multimodal feature fusion module comprises two self-attention transformer~(sat) blocks, two cross-attention transformer~(cat) blocks, and two concatenation operation~(concat), as shown in Algorithm~\ref{algorithm2}. 
	
The sat blocks employ standard transformers, which are characterized by multihead self-attention~(MSA) and multilayer perceptrons~(MLP), to enhance the feature patch embeddings~\citep{dosovitskiy2020image}:
\begin{equation}
\tilde{X} = {\rm MSA}\left({\rm LN}(X)\right)+X,
\label{eq11_1}
\end{equation}
\begin{equation}
X = {\rm MLP}\left(\tilde{X}\right)+\tilde{X}.
\label{eq11_2}
\end{equation}
Here, $\rm LN$ represents the Layer Normalization operation, and $X$ denotes the input patch embeddings. In each cat block, a modified cross-attention~(CA) mechanism is utilized to replace the self-attention in MSA:
\begin{equation}
{\rm CA} = {\rm softmax} \left ( \frac{X D^{T}}{\sqrt{D_{\rm dim}} }  \right ) \times D ,
\label{eq11}
\end{equation}
where $X^{N\times D_{\rm dim}}$ denotes the enhanced embeddings and $D^{2N\times D_{\rm dim}}$ represents the fusion embeddings. In a certain sense, the proposed CA mechanism facilitates the extensive target-specific modeling of visual cues across different domains. Using the MLP and multihead cross-attention~(MCA) refined with ca mechanism, the entire process of cat block can be described as follows:
\begin{equation}
\tilde{X} = {\rm MCA}\left({\rm LN}(X)\right)+X,
\label{eq11_3}
\end{equation}
\begin{equation}
X = {\rm MLP}\left(\tilde{X}\right)+\tilde{X}.
\label{eq11_4}
\end{equation}
Similar to a general transformer encoder, our feature fusion module also enables repetitive embedding processes. 

To date, the fusion patch embeddings of low-level $T_{\rm l,embed}^{i}$ and high-level $T_{\rm h,embed}^{i}$ are obtained:
\begin{equation}
T_{\rm l,embed}^{i} = \text{Algorithm2}\left ( F_{\rm l,embed}^{i}, G_{\rm l,embed}^{i} \right ) ,
\label{eq11_5}
\end{equation}
\begin{equation}
T_{\rm h,embed}^{i} = \text{Algorithm2}\left ( F_{\rm h,embed}^{i}, G_{\rm h,embed}^{i} \right ) .
\label{eq11_6}
\end{equation}

\begin{table*}[t]\footnotesize
	\caption{Details of model configurations for different datasets. In the MMFE module, for example, C64k11s4p5 signifies a convolutional layer with 64 output channels, kernel size 11, stride 4, and padding 5. BN represents batch normalization. In the TFF module, parameters include the resolution of feature patches $p$, the embeddings dimensionality $D_{dim}$ and the number of heads in MSA and MCA.}
	\centering
	\setlength{\tabcolsep}{6.5mm}{
\begin{tabular}{l|cc|ccc}
	\hline
	\multirow{2}{*}{Datasets} & \multicolumn{2}{c|}{MMFE}                                                                                                 & \multicolumn{3}{c}{TFF~(\#Block = 2)}                                                      \\ \cline{2-6} 
	& \multicolumn{1}{c|}{Convl\_x}                                                                             & Convh\_x      & \multicolumn{1}{c|}{$p$} & \multicolumn{1}{c|}{$D_{\text{dim}}$}    & \multicolumn{1}{c}{\#head}  \\ \hline
	FE108                     & \multicolumn{1}{c|}{\begin{tabular}[c]{@{}c@{}}C64k11s4p5-BN-C128k5s2p2-BN\\ -C128k3s1p1-BN\end{tabular}} & C256k3s2p1-BN & \multicolumn{1}{c|}{4} & \multicolumn{1}{c|}{512} & \multicolumn{1}{c}{2}         \\
	VisEvent                  & \multicolumn{1}{c|}{C64k11s4p5-C128k5s2p2}                                                                & C256k3s2p1    & \multicolumn{1}{c|}{4} & \multicolumn{1}{c|}{512} & \multicolumn{1}{c}{2}         \\
	COESOT                    & \multicolumn{1}{c|}{C64k11s4p5-C128k5s2p2}                                                                & C256k3s2p1    & \multicolumn{1}{c|}{4} & \multicolumn{1}{c|}{512} & \multicolumn{1}{c}{2}         \\ \hline
\end{tabular}}
\label{table0}
\end{table*}

\begin{table*}[t]\footnotesize
	\caption{Comparison among the existing large-scale frame-event-based datasets for object tracking. The \# symbol represents the number of corresponding items.}
	\centering
	\setlength{\tabcolsep}{4.7mm}{
		\begin{tabular}{l|ccccccc}
			\hline
			Datasets  & \multicolumn{1}{l}{Year} & \multicolumn{1}{l}{\#Videos} & \#Train/Test & \multicolumn{1}{l}{\#Frames} &  Resolution  & \multicolumn{1}{l}{\#Attributes} & Device \\ \hline
			FE108     &           2021           &             108              &    76/32     &            200157            & 346$\times$260 &                4                 & DAVIS346 \\
			VisEvent  &           2021           &             746              &   445/301    &            323220            & 346$\times$260 &                17                & DAVIS346 \\
			COESOT    &           2022           &             1354             &   827/527    &            466833            & 346$\times$260 &                17                & DAVIS346 \\ \hline
	\end{tabular}}
	\label{table1}
\end{table*}

\subsubsection{Multimodal Feature Decoder}
To provide inputs for the tracking heads, we design a multimodal feature decoder to convert the embeddings into fusion-level feature maps~(given in Algorithm~\ref{algorithm3}). In our decoder, the embeddings in each dimension are projected into a new feature space using a linear layer, and its distribution is adjusted by layer normalization. To date, multimodal feature modeling has been accomplished, yielding fusion-level feature maps $T_{\rm l}^{i}$ and $T_{\rm h}^{i}$.

\subsection{Heads and Loss}
For the tracking heads, namely, the regressor and classifier, we employ the target estimation network from ATOM~\citep{danelljan2019atom} and the classifier from DiMP~\citep{bhat2019learning}, respectively. The regressor, characterized by modulation~($IoU_{\rm mod}$) and prediction~($IoU_{\rm pre}$) blocks, takes as low-level and high-level feature maps~($T_{\rm l}^{i}$ and $T_{\rm h}^{i}$, respectively) as inputs to estimate $IoU^{i}$. Mathematically, the computational procedure can be expressed as follows:	 
\begin{equation}
v_{\rm l},v_{\rm h}=IoU_{\rm mod}(T_{\rm l,t}^{i}, T_{\rm h,t}^{i}, B_{\rm t}),
\label{add1}
\end{equation}
\begin{equation}
IoU^{i}=IoU_{\rm pre}(T_{\rm l,s}^{i}, T_{\rm h,s}^{i}, B_{\rm s},v_{\rm l},v_{\rm h}).
\label{add2}
\end{equation}
Here, subscript $\rm t$ and $\rm s$ denote template and search frame, respectively. The symbol $B$ represents the target bounding box. On the other hand, the classifier utilizes $T_{\rm h}^{i}$ to predict a confidence score $s^{i}$ for the target as follows:
\begin{equation}
s^{i}=Classifier(T_{h,t}^{i}, T_{h,s}^{i}, B_{t}).
\label{add3}
\end{equation}
Notably, the discriminative filter generated in the classifier can be learned online.

The loss function $L_{{\rm total}}$ for offline training is defined as follows:
\begin{equation}
L_{{\rm total}}=\beta L_{{\rm cls}} + L_{{\rm reg}},
\label{eq12}
\end{equation}
with
\begin{equation}
L_{{\rm cls}} = \frac{1}{I} \sum_{i=1}^{I}   \zeta^{2}(s^{i},s_{\rm gt}^{i}),
\label{eq13}
\end{equation}
\begin{equation}
\zeta (s^{i}, s_{\rm gt}^{i})=\begin{cases}
	s^{i}-s_{\rm gt}^{i},  &if\ s_{\rm gt}^{i}  >  0.05\\
	{\rm max}(0,s^{i}),    &if\ s_{\rm gt}^{i}\le  0.05\\
\end{cases},
\label{eq14} 
\end{equation}
\begin{equation}
L_{{\rm reg}} = \frac{1}{I} \sum_{i=1}^{I} (IoU^{i}-IoU_{\rm gt}^{i})^{2},
\label{eq15}
\end{equation}
where $s_{\rm gt}^{i}$ is a Gaussian label generated according to the corresponding ground truth $IoU_{\rm gt}^{i}$. The losses of the classifier $L_{{\rm cls}}$ and regressor $L_{{\rm reg}}$ represent the mean squared error determined on $I$ samples. The constant coefficient $\beta$ is used to balance the weight between two heads.

\begin{figure}[t]
\centering
\includegraphics[width=1.0\columnwidth]{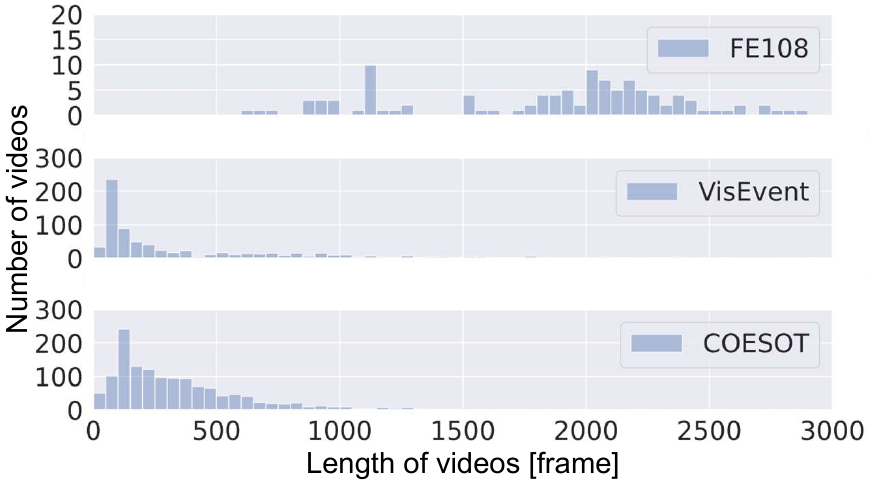}
\caption{The video length distribution across the datasets, with a histogram interval of 50 frames and an upper bound of 3000 frames in the statistics.}
\label{Figure3}
\end{figure}

\section{Experiments}
In this section, we begin by providing a comprehensive overview of our experimental settings, encompassing the utilized datasets, evaluation metrics, and preprocessing steps. Subsequently, we present a detailed performance comparison between our proposed MMHT model and other state-of-the-art models on diverse benchmark datasets. Furthermore, we conduct rigorous ablation studies to demonstrate the indispensability of multimodal tracking, the hybrid backbones, and the proposed fusion components. To achieve enhanced comprehension, representative figures are given to provide qualitative visualizations.

\subsection{Experimental Settings}
\begin{table*}[t]\footnotesize
	\caption{Comparison among the state-of-the-art performance metrics, including PR, SR, OP50, OP75 and FPS, achieved on the FE108, COESOT and VisEvent. The average results obtained by the MMHT model are presented as means $\pm$ standard deviations. The best results are emphasized in bold. The annotations of the CEUT model indicate different data processing approaches. Note: The ATOM and PrDiMP models were originally proposed in Ref.~\citep{danelljan2019atom} and Ref.~\citep{danelljan2020probabilistic}, respectively. }
	\centering
	\setlength{\tabcolsep}{4.4mm}{
		\begin{tabular}{c|c|ccccc}
			\hline
			Method                               &           Fusion level            &                PR[\rm$\%$]                 &                SR[\rm$\%$]                 &               OP50[\rm$\%$]                &               OP75[\rm$\%$]             &FPS   \\ \hline
			\multicolumn{6}{c}{FE108}                                                                                                                                  \\ \hline
			\multicolumn{1}{c|}{ATOM+Event~\citep{zhang2021object}}       &            data-level             &         \multicolumn{1}{c}{81.80}          &         \multicolumn{1}{c}{55.50}          &         \multicolumn{1}{c}{70.00}          &         \multicolumn{1}{c}{27.40}         &- \\
			\multicolumn{1}{c|}{PrDiMP+Event~\citep{zhang2021object}}      &            data-level             &         \multicolumn{1}{c}{87.70}          &         \multicolumn{1}{c}{59.00}          &         \multicolumn{1}{c}{74.40}          &         \multicolumn{1}{c}{29.80}         &- \\
			\multicolumn{1}{c|}{CEUT$_{1}$~\citep{tang2022revisiting}}      &         unified backbone          &         \multicolumn{1}{c}{84.46}          &         \multicolumn{1}{c}{55.58}          &          \multicolumn{1}{c}{\--}           &          \multicolumn{1}{c}{\--}        &-  \\
			\multicolumn{1}{c|}{RT-MDNet~\citep{wang2021visevent}}         &           feature-level           &         \multicolumn{1}{c}{56.40}          &         \multicolumn{1}{c}{35.90}          &          \multicolumn{1}{c}{\--}           &          \multicolumn{1}{c}{\--}         &14  \\
			\multicolumn{1}{c|}{CDFI~\citep{zhang2021object}}          &           feature-level           &         \multicolumn{1}{c}{92.40}          &     \multicolumn{1}{c}{\textbf{63.40}}     &         \multicolumn{1}{c}{81.30}          &     \multicolumn{1}{c}{\textbf{34.40}}   & 30 \\
			\multicolumn{1}{c|}{MMHT~(Ours)}                  &           feature-level           & \multicolumn{1}{c}{\textbf{93.62$\pm$.33}} &     \multicolumn{1}{c}{62.97$\pm$.11}      & \multicolumn{1}{c}{\textbf{81.68$\pm$.26}} &     \multicolumn{1}{c}{29.92$\pm$.15}     &17 \\ \hline
			\multicolumn{6}{c}{COESOT}                                                                                                                                 \\ \hline
			\multicolumn{1}{c|}{OSTrack~\citep{ye2022joint}}           &            data-level             &         \multicolumn{1}{c}{66.60}          &         \multicolumn{1}{c}{59.00}          &          \multicolumn{1}{c}{\--}           &          \multicolumn{1}{c}{\--}          &105 \\
			\multicolumn{1}{c|}{SiamR-CNN+Event~\citep{tang2022revisiting}}      &            data-level             &         \multicolumn{1}{c}{67.50}          &         \multicolumn{1}{c}{60.90}          &          \multicolumn{1}{c}{\--}           &          \multicolumn{1}{c}{\--}          &5 \\
			\multicolumn{1}{c|}{KeepTrack+Event~\citep{tang2022revisiting}}     &            data-level             &         \multicolumn{1}{c}{66.10}          &         \multicolumn{1}{c}{59.60}          &          \multicolumn{1}{c}{\--}           &          \multicolumn{1}{c}{\--}          &18 \\
			\multicolumn{1}{c|}{CEUT$_{1}$~\citep{tang2022revisiting}}      &         unified backbone          &         \multicolumn{1}{c}{70.50}          &         \multicolumn{1}{c}{62.00}          &          \multicolumn{1}{c}{\--}           &          \multicolumn{1}{c}{\--}          & 75 \\
			\multicolumn{1}{c|}{CEUT$_{2}$~\citep{tang2022revisiting}}      &         unified backbone          &         \multicolumn{1}{c}{68.60}          &         \multicolumn{1}{c}{60.40}          &          \multicolumn{1}{c}{\--}           &          \multicolumn{1}{c}{\--}          &- \\
			\multicolumn{1}{c|}{MDNet-MF~\citep{tang2022revisiting}}      &         feature-level          &         \multicolumn{1}{c}{64.70}          &         \multicolumn{1}{c}{56.30}          &          \multicolumn{1}{c}{\--}           &          \multicolumn{1}{c}{\--}          & 14 \\
			\multicolumn{1}{c|}{MMHT~(Ours)}                  &           feature-level           & \multicolumn{1}{c}{\textbf{74.03$\pm$.20}} & \multicolumn{1}{c}{\textbf{65.81$\pm$.12}} & \multicolumn{1}{c}{\textbf{77.64$\pm$.12}} & \multicolumn{1}{c}{\textbf{56.97$\pm$.14}}  &19\\ \hline
			\multicolumn{6}{c}{VisEvent}                                                                                                                                \\ \hline
			\multicolumn{1}{c|}{CEUT$_{1}$~\citep{tang2022revisiting}}      &         unified backbone          &         \multicolumn{1}{c}{69.06}          &         \multicolumn{1}{c}{53.12}          &          \multicolumn{1}{c}{\--}           &          \multicolumn{1}{c}{\--}        & -   \\
			\multicolumn{1}{c|}{ViPT~\citep{zhu2023visual}}           &         unified backbone          &         \multicolumn{1}{c}{75.80}          &         \multicolumn{1}{c}{59.20}          &          \multicolumn{1}{c}{\--}           &          \multicolumn{1}{c}{\--}        &-   \\
			\multicolumn{1}{c|}{Un-Track~\citep{wu2023single}} & unified backbone &     \multicolumn{1}{c}{\textbf{76.30}}     &     \multicolumn{1}{c}{\textbf{59.70}}     &          \multicolumn{1}{c}{\--}           &          \multicolumn{1}{c}{\--}         &-  \\
			\multicolumn{1}{c|}{ProTrack~\citep{yang2022prompting}}       &           prompt-based            &         \multicolumn{1}{c}{61.70}          &         \multicolumn{1}{c}{47.40}          &          \multicolumn{1}{c}{\--}           &          \multicolumn{1}{c}{\--}          &- \\
			\multicolumn{1}{c|}{SiamFC~\citep{qiaog}}              &           feature-level           &         \multicolumn{1}{c}{52.30}          &         \multicolumn{1}{c}{35.00}          &          \multicolumn{1}{c}{\--}           &          \multicolumn{1}{c}{\--}          &- \\
			\multicolumn{1}{c|}{AFNet~\citep{zhang2023frame}}          &           feature-level           &         \multicolumn{1}{c}{59.30}          &         \multicolumn{1}{c}{44.50}          &          \multicolumn{1}{c}{\--}           &          \multicolumn{1}{c}{\--}          &- \\
			\multicolumn{1}{c|}{MMHT~(Ours)}                  &           feature-level           &     \multicolumn{1}{c}{73.26$\pm$.11}      &     \multicolumn{1}{c}{55.10$\pm$.16}      &     \multicolumn{1}{c}{65.94$\pm$.20}      &     \multicolumn{1}{c}{42.78$\pm$.14}      &21\\ \hline
	\end{tabular}}
	\label{table2}
\end{table*}

\begin{table*}[t]\footnotesize
\caption{Analysis of the performance achieved by trackers trained with single-modal and multimodal data. The left section showcases the comprehensive evaluation results, while the right section provides specific comparisons pertaining to typical diverse attributes.}
\centering
\setlength{\tabcolsep}{1.0mm}{
	\begin{tabular}{cc|cc|cc|cc|cc|cc}
		\hline
		\multirow{6}{*}{\rotatebox{90}{FE108}}         & \multirow{3}{*}{Modality} &    \multicolumn{2}{c|}{\multirow{2}{*}{ALL}}    &                                                                                     \multicolumn{8}{c}{Attributes}                                                                                       \\ \cline{5-12}
		&                           &              \multicolumn{2}{c|}{}              &             \multicolumn{2}{c|}{LL}              &            \multicolumn{2}{c|}{HDR}             &             \multicolumn{2}{c|}{FWB}             &             \multicolumn{2}{c}{FNB}             \\ \cline{3-12}
		&                           &      PR[\rm$\%$]       &      SR[\rm$\%$]       &       PR[\rm$\%$]       &      SR[\rm$\%$]       &      PR[\rm$\%$]       &      SR[\rm$\%$]       &       PR[\rm$\%$]       &      SR[\rm$\%$]       &      PR[\rm$\%$]       &      SR[\rm$\%$]                            \\ \cline{2-12}
		&           Frame           &     72.48$\pm$.44      &     48.48$\pm$.25      &     43.59$\pm$1.07      &     29.09$\pm$.64      &     62.11$\pm$.83      &     40.56$\pm$.46      &      99.72$\pm$.04      &     68.22$\pm$.05      &     94.78$\pm$.35      & \textbf{62.22$\pm$.22}                      \\
		&           Event           &     84.18$\pm$.50      &     55.44$\pm$.30      &      96.19$\pm$.20      &     66.00$\pm$.21      &     81.84$\pm$.27      &     52.06$\pm$.17      & \textbf{100.00$\pm$.00} &     69.83$\pm$.05      &     62.51$\pm$2.10     &     35.64$\pm$1.07                          \\
		&          Fusion           & \textbf{93.62$\pm$.33} & \textbf{62.97$\pm$.11} & \textbf{96.37$\pm$1.20} & \textbf{66.02$\pm$.73} & \textbf{89.43$\pm$.54} & \textbf{58.13$\pm$.23} &      99.80$\pm$.02      & \textbf{71.34$\pm$.06} & \textbf{95.53$\pm$.36} &     61.93$\pm$.18                           \\ \hline
		\multirow{6}{*}{\rotatebox{90}{COESOT}}         & \multirow{3}{*}{Modality} &    \multicolumn{2}{c|}{\multirow{2}{*}{ALL}}    &                                                                                     \multicolumn{8}{c}{Attributes}                                                                                       \\ \cline{5-12}
		&                           &              \multicolumn{2}{c|}{}              &             \multicolumn{2}{c|}{BOM}             &             \multicolumn{2}{c|}{BC}             &             \multicolumn{2}{c|}{SV}              &             \multicolumn{2}{c}{VC}             \\ \cline{3-12}
		&                           &      PR[\rm$\%$]       &      SR[\rm$\%$]       &       PR[\rm$\%$]       &      SR[\rm$\%$]       &      PR[\rm$\%$]       &      SR[\rm$\%$]       &       PR[\rm$\%$]       &      SR[\rm$\%$]       &      PR[\rm$\%$]       &      SR[\rm$\%$]                             \\ \cline{2-12}
		&           Frame           &     66.16$\pm$.20      &     62.51$\pm$.08      &      62.84$\pm$.21      &     60.65$\pm$.08      &     50.82$\pm$.36      &     48.51$\pm$.18      &      68.08$\pm$.50      &     64.92$\pm$.38      &     62.97$\pm$.68      &     60.89$\pm$.64                            \\
		&           Event           &     52.34$\pm$.16      &     52.98$\pm$.12      &      50.03$\pm$.27      &     52.31$\pm$.20      &     43.66$\pm$.26      &     43.84$\pm$.23      &      47.87$\pm$.35      &     50.07$\pm$.14      &     45.50$\pm$.65      &     50.25$\pm$.53                            \\
		&          Fusion           & \textbf{74.03$\pm$.20} & \textbf{65.81$\pm$.12} & \textbf{73.20$\pm$.25}  & \textbf{65.52$\pm$.13} & \textbf{67.41$\pm$.41} & \textbf{57.53$\pm$.21} & \textbf{71.51$\pm$.29}  & \textbf{65.38$\pm$.18} & \textbf{67.20$\pm$.24} & \textbf{64.61$\pm$.15}                       \\ \hline
		\multirow{6}{*}{\rotatebox{90}{VisEvent}} & \multirow{3}{*}{Modality} &    \multicolumn{2}{c|}{\multirow{2}{*}{ALL}}    &                                                                                     \multicolumn{8}{c}{Attributes}                                                                                      \\ 
		\cline{5-12}
		\cline{5-12}
		&                           & \multicolumn{2}{c|}{}                           & \multicolumn{2}{c|}{BOM}                         & \multicolumn{2}{c|}{BC}                         & \multicolumn{2}{c|}{SV}                          &                     
		\multicolumn{2}{c}{VC}             \\ \cline{3-12}
		&                           & PR[\rm$\%$]                 & SR[\rm$\%$]                 & PR[\rm$\%$]                  & SR[\rm$\%$]                 & PR[\rm$\%$]                 & SR[\rm$\%$]                 & PR[\rm$\%$]                  & SR[\rm$\%$]                 & PR[\rm$\%$]                 & SR[\rm$\%$]                  \\ 
		\cline{2-12}
		& Frame                     & 69.26$\pm$.10          & 52.84$\pm$.15          & 65.27$\pm$.20           & 49.73$\pm$.14          & 65.23$\pm$.20          & 49.10$\pm$.22          & 62.30$\pm$.14           & 48.07$\pm$.19          & 58.66$\pm$.09          & 46.56$\pm$.26           \\
		& Event                     & 62.70$\pm$.14          & 48.47$\pm$.22          & 57.69$\pm$.23           & 44.74$\pm$.04          & 57.36$\pm$.19          & 44.42$\pm$.22          & 58.73$\pm$.14           & 45.72$\pm$.16          & 53.59$\pm$.26          & 42.72$\pm$.17           \\
		& Fusion                    & \textbf{73.26$\pm$.25} & \textbf{55.10$\pm$.15} & \textbf{69.75$\pm$.16}  & \textbf{52.42$\pm$.08} & \textbf{70.10$\pm$.23} & \textbf{51.83$\pm$.08} & \textbf{67.33$\pm$.15}  & \textbf{50.58$\pm$.22} & \textbf{68.40$\pm$.11} & \textbf{52.44$\pm$.11}  \\                      \hline
\end{tabular}}
\label{table3}
\end{table*}

\subsubsection{Datasets}
In our experiments, the FE108~\citep{zhang2021object}, VisEvent~\citep{wang2021visevent} and COESOT~\citep{tang2022revisiting} datasets, captured in real scenes using a DAVIS346 event camera, are utilized to train and test our trackers. The DAVIS346 camera enables the simultaneous acquisition of aligned frame-event-based data with a spatial resolution of 346$\times$260. 

As the details of the datasets shown in Tab.~\ref{table1}, the FE108 dataset consists of 108 annotated videos, with 72 videos used for training and 32 videos employed for testing. These videos are categorized based on four attributes: low light~(LL), high dynamic range~(HDR), fast motion with motion blur~(FWB), and fast motion without motion blur~(FNB). The VisEvent dataset, which is also used in our work, includes 746 annotated videos, with 445 videos for training and 301 videos for testing. The COESOT dataset comprises 1354 annotated videos, with 827 videos for training and 527 videos for testing. Notably, due to missing raw data and annotations, we refilter the VisEvent dataset. Both datasets cover 17 representative attributes. However, our work focuses on four specific attributes: background object motion~(BOM), background clutter~(BC), scale variation~(SV), and viewpoint change~(VC). 

Additionally, we analyze the video lengths distributions of the three datasets. As depicted in Fig.~\ref{Figure3}, the FE108 dataset has the fewest number of videos but the longest video length. On the other hand, the VisEvent and COESOT datasets exhibit centralized video length distributions, with videos mostly within 500 frames. These datasets provide a comprehensive understanding of the performance achieved by the model in both long- and short-term tracking scenarios.

\subsubsection{Evaluation Metrics}
To validate the performance of our trackers, we plot the precision and success curves of our testing results. The precision curve illustrates the percentage of frames where the center distance between the predicted and ground-truth bounding boxes falls within a specified threshold. The success curve focuses on the frames where the overlap between the predicted and ground-truth bounding boxes exceeds a given threshold. In our study, we employ several quantitative metrics for evaluation purposes: the precision rate~(PR) measured with 20 pixels as the threshold; the success rate~(SR) represented by the area under the success curve; and two overlap precision rates~(OP50 and OP75), indicating the success rates achieved at overlap levels of $0.50$ and $0.75$, respectively.

\subsubsection{Implementation Details}
The MMHT model is implemented using PyTorch and executed on a workstation equipped with NVIDIA A100 GPUs. The training process of our trackers consists of 50 epochs, with a batch size of 20 and the adaptive moment estimation~(Adam) optimizer~\citep{kingma2014adam} with its default parameters. For the FE108 dataset, the initial learning rates of the hybrid backbones and the other components are set to $0.0001$ and $0.001$, respectively. For the VisEvent and COESOT datasets, the ANN backbone has an initial learning rate of $0.00001$, but the other parameters are set to $0.001$. In the SNN backbones, all trainable spike thresholds $u_{\rm th}$ are initialized to $1.0$, and the membrane decay constant $\alpha$ is fixed at $0.7$. All the learning rates follow the exponential decay process with a factor of $0.9$. All trackers are tested 5 times, and the reported results are averaged. The code of our implementation will be available at https://github.com/GuoLab-UESTC after this manuscript is accepted for publication.

\begin{figure}[t]
\centering
\includegraphics[width=1.0\columnwidth]{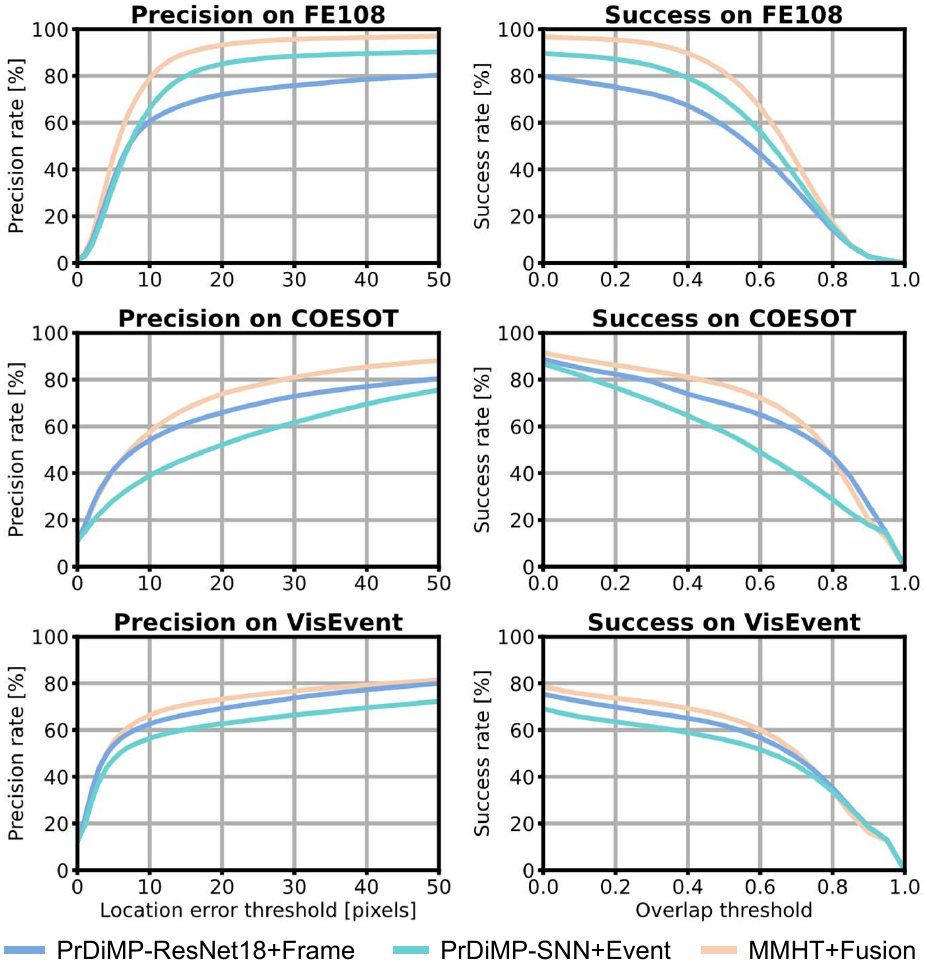} 
\caption{The precision and success curves yielded by trackers trained with different modalities.}
\label{Figure4}
\end{figure}

\begin{figure*}[t]
\centering
\includegraphics[width=2.0\columnwidth]{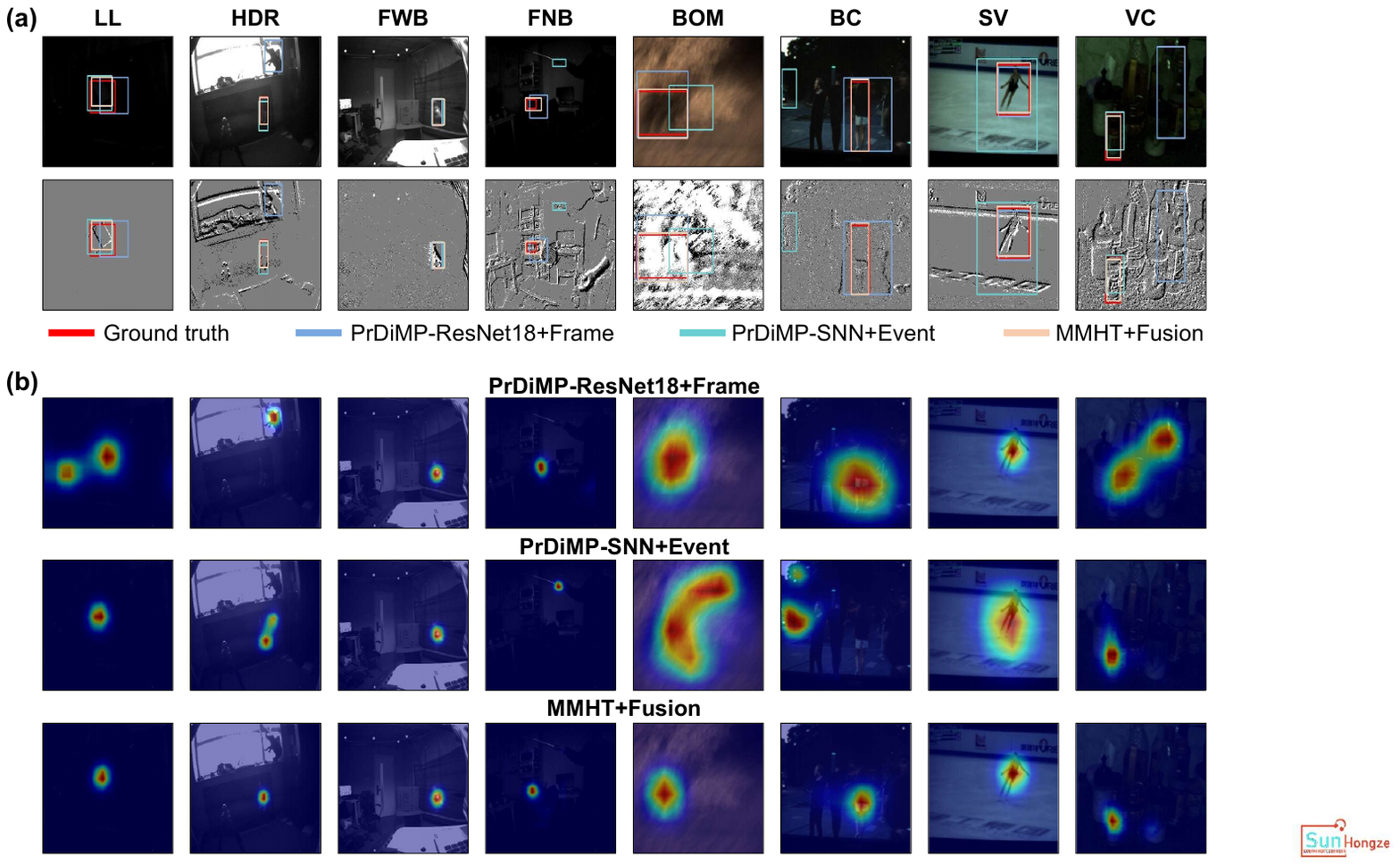}
\caption{Visualization of the results produced by trackers trained using diverse modalities. (a) Tracking results obtained from trackers trained with various modalities. The predicted bounding boxes generated by the trackers are visually compared with the ground truth bounding boxes of the input images obtained from two modalities. (b) Corresponding response maps of different trackers. The response  intensity progresses from green to red, indicating an increasing response level.}
\label{Figure5}
\end{figure*}

\subsection{Comparison with the State-of-the-Art Methods}
To validate the effectiveness of our proposed MMHT, we conduct a comparative analysis with other state-of-the-art trackers~\citep{zhang2021object,tang2022revisiting, wang2021visevent, ye2022joint, zhang2023frame, yang2022prompting, qiaog} on the FE108, COESOT, and VisEvent datasets. According to their fusion levels, these trackers can be classified into various categories: data-level fusion trackers~(where data from multiple  sources are combined at the input layer), feature-level fusion trackers~(where features are separately extracted from different modalities and combined to create a unified representation for tracking purposes), unified backbone fusion trackers~(which utilize a single backbone network to process data from multiple modalities), and prompt-based fusion trackers~(where multiple modalities serve as a prompt guide for reliable visible image tracking).

As shown in Tab.~\ref{table2}, the MMHT outperforms the other methods on both the FE108 and COESOT datasets in terms of a majority of the utilized metrics. Notably, on the FE108 dataset, the MMHT achieves a $1.22\%$ PR improvement and a $0.38\%$ OP50 improvement over the previous best method. On the COESOT dataset, our model demonstrates PR and SR improvements of $3.53\%$ and $3.81\%$, respectively. To our knowledge, our work is the first to publish OP50 and OP75 results obtained on the COESOT dataset. On the VisEvent dataset, our MMHT model yields slightly lower PR and SR results than those of the current state-of-the-art method~(i.e., see ViPT and Un-Track in Tab.~\ref{table2}). To a certain extent, this might be due to the partial absence of raw data in the VisEvent dataset, causing different models to use different numbers of training and test samples in the experiments.  Furthermore, by comparing the distributions of video lengths reported in the previous study~\citep{wang2021visevent}, the absent data primarily concentrates in long videos exceeding 1000 frames. Considering the substantial improvement of MMHT on datasets FE108 and COESOT with a higher proportion of long videos, we posit that the absence of long videos may also impact the evaluation performance of MMHT. Remarkably, when compared to the existing state-of-the-art feature-level fusion trackers, we find that the MMHT achieves notable PR advancements~(with a substantial increase of $13.96\%$) and a significant SR improvement~(with a boost of $10.60\%$).

Tracking speed, commonly quantified in frames per second (FPS), is a crucial metric for evaluating tracker performance in real applications. As illustrated in Tab.~\ref{table2}, the tracking speeds of the MMHT model (17, 19, and 21 FPS on FE108, COESOT, and VisEvent, respectively) fall within the mid-range and are just lower than those of specific models~(CDFI~30 FPS, OSTrack~105 FPS, and CEUT~75 FPS). It is evident that such intermediate performance of the MMHT model in tracking speed is attributable to the increased computational complexity introduced by the two-stream hybrid strategy. However, by considering the significant accuracy improvements of the model on different datasets, we posit that the tracking speeds of MMHT are still acceptable for real applications.

Overall, these observations demonstrate the superiority of our proposed MMHT model, which can exhibit competitive performance in comparison with that of the previously developed state-of-the-art methods on various benchmark datasets.

\subsection{Ablation Studies}
\subsubsection{Analysis of the Visual Modality}
To further illustrate the benefits of employing multimodal data for object tracking, we conduct an ablation analysis on trackers trained exclusively on single-modal data. Specifically, we retain corresponding input and backbone modules in MMHT tailored to the utilized modality. The TFF module is removed, while the tracking heads remain unaltered. In these single-modal trackers, denoted as PrDiMP-SNN+Event and PrDiMP-ResNet18+Frame, tracking heads directly receive both low-level and high-level features to generate predictions. The precision and success curves are depicted in Fig.~\ref{Figure4}. In summary, our multimodal MMHT models demonstrate significantly wider performance margins than those of their single-modal counterparts across different datasets.

Quantitative comparison: The precise results are presented in Tab.~\ref{table3}. When employing single-modal tracking on FE108, the event modality demonstrates significant advantages over the frame modality. However, the incorporation of multimodal data still yields improvements of $9.44\%$ and $7.53\%$ in terms of the PR and SR metrics, respectively. On the more challenging COESOT and VisEvent datasets, the frame modality exhibits a notable advantage. Nevertheless, with the utilization of multimodal data, we observe a substantial PR and SR increase of $7.87\%$ and $3.30\%$ on the COESOT dataset and $4.00\%$ and $2.26\%$ on the VisEvent dataset, respectively. These results affirm the outstanding multimodal tracking performance of the proposed approach.

Attribute-based comparison: We roughly divide several typical attributes into two categories: 1. environment-oriented attributes, including LL, HDR, BOM, and BC, and 2. object-oriented attributes including FWB, FNB, SV, and VC. Specifically, in challenging scenarios influenced by environmental factors, the event modality exhibits pronounced advantages in scenarios with LL and HDR. This advantage stems from the higher dynamic ranges of the photodetector used by the event camera. However, in BOM and BC scenarios, the event modality may be susceptible to significant discriminative filter drift due to the absence of texture features, rendering it less effective than the frame modality. In challenging scenarios caused by the target objects, the frame modality excels in cases with FNB, SV, and VC. We posit that these scenarios may disrupt the temporal features of  the event data, whereas the abundant spatial information in the frame modality remains relatively unaffected. However, in the FWB case, both single-modal trackers achieve nearly perfect scores in terms of the PR and SR metrics. Nevertheless, the multimodal MMHT still demonstrates superiority across most attributes.

\begin{table*}[t]\footnotesize
\caption{The MAC and AC operations in both the ANN and SNN backbones. $E_{\text{MAC}}$ and $E_{\text{AC}}$ represent the empirical energy consumption values. $K_{n}$ denotes the size of the convolutional kernels in the $n$-th layer. $H$, $W$, and $C$ refer to the height, width, and channel dimensions of the feature maps, respectively. $FR$ signifies the average firing rate of SNNs~\citep{chen2022adaptive}.}
\centering
\begin{tabular}{p{1.3cm}<{\centering}|p{3.0cm}<{\centering}|p{5.0cm}<{\centering}|p{5.0cm}<{\centering}}
	\hline
	\multirow{2}{*}{OPs} & \multirow{2}{*}{Power consumption} &                                                                \multicolumn{2}{c}{The number of OPs within backbones}                                                                 \\ \cline{3-4}
	&                                    & ANN                                                                              & SNN                                                                                                \\ \hline
	MAC                        & $E_{\rm{MAC}}=4.6pJ$               & $MAC_{\rm{ANN}}=\sum_{n=1}^{N} K_n^2\cdot C_{n-1}\cdot H_n \cdot W_n \cdot C_n $ & $MAC_{\rm{SNN}} = N \cdot K_1^2\cdot C_0\cdot H_1 \cdot W_1 \cdot C_1 $                            \\ 
	AC                         & $E_{\rm{AC}}=0.9pJ$                & $AC_{\rm{ANN}}=0$                                                                & $AC_{\rm{SNN}}=N \cdot \sum_{n=2}^{N}FR_n \cdot  K_n^2\cdot C_{n-1}\cdot H_n \cdot W_n \cdot C_n $ \\ \hline
\end{tabular}
\label{table4}
\end{table*}

\begin{table}[t]\footnotesize
	\caption{Performance analysis of various backbones with respect to event feature extraction.}
	\centering
	\setlength{\tabcolsep}{2.5mm}{
		\begin{tabular}{cc|c|cc}
			\hline
			                   \multicolumn{2}{c|}{Modality}                    &   Model    &      PR[\rm$\%$]       &      SR[\rm$\%$]       \\ \hline
			 \multirow{4}{*}{\rotatebox{90}{FE108}}   & \multirow{2}{*}{Event}  & PrDiMP-ANN &     79.91$\pm$.34      &     50.63$\pm$.19      \\
			                                          &                         & PrDiMP-SNN & \textbf{84.18$\pm$.50} & \textbf{55.44$\pm$.30} \\ \cline{2-5}
			                                          & \multirow{2}{*}{Fusion} & MMHT-ANN &     68.25$\pm$.29      &     43.99$\pm$.21      \\
			                                          &                         &    MMHT    & \textbf{93.62$\pm$.33} & \textbf{62.97$\pm$.11} \\ \hline
			 \multirow{4}{*}{\rotatebox{90}{COESOT}}  & \multirow{2}{*}{Event}  & PrDiMP-ANN &     50.37$\pm$.08      &     52.78$\pm$.11      \\
			                                          &                         & PrDiMP-SNN & \textbf{52.34$\pm$.16} & \textbf{52.98$\pm$.12} \\ \cline{2-5}
			                                          & \multirow{2}{*}{Fusion} & MMHT-ANN &     70.95$\pm$.29      &     62.55$\pm$.18      \\
			                                          &                         &    MMHT    & \textbf{74.03$\pm$.20} & \textbf{65.81$\pm$.12} \\ \hline
			\multirow{4}{*}{\rotatebox{90}{VisEvent}} & \multirow{2}{*}{Event}  & PrDiMP-ANN &     45.48$\pm$.51      &     35.89$\pm$.37      \\
			                                          &                         & PrDiMP-SNN & \textbf{62.70$\pm$.14} & \textbf{48.47$\pm$.22} \\ \cline{2-5}
			                                          & \multirow{2}{*}{Fusion} & MMHT-ANN &     65.24$\pm$.33      &     50.95$\pm$.18      \\
			                                          &                         &    MMHT    & \textbf{73.26$\pm$.25} & \textbf{55.10$\pm$.15} \\ \hline
		\end{tabular}}
	\label{table5}
\end{table}

Qualitative visualizations: To achieve intuitive comprehension, we randomly select a representative sample from each of the eight attributes for visual analysis purposes~(4 from FE108 and 4 from COESOT and VisEvent). As illustrated in Fig.~\ref{Figure5}, we plot the tracking results~(Fig.~\ref{Figure5}(a)) and corresponding response maps~(Fig.~\ref{Figure5}(b)) yielded by trackers trained with diverse modalities. The visualized results align well with the quantitative findings. Except for the relatively simple FWB scenario, in which nearly no discernible differences are observed among the  performances of all models, the trackers trained with a single modality exhibit instances of identification errors~(HDR, FNB, BC and VC) or response drift~(LL, BOM and SV) across most attributes. In contrast, the MMHT models exhibit more precise responses to objects, yielding more accurate bounding box predictions. The visualization analysis further substantiates the exceptional robustness and superiority of MMHT models in terms of achieving effective object tracking across a spectrum of challenging scenarios.

\begin{figure}[t]
\centering
\includegraphics[width=1.0\columnwidth]{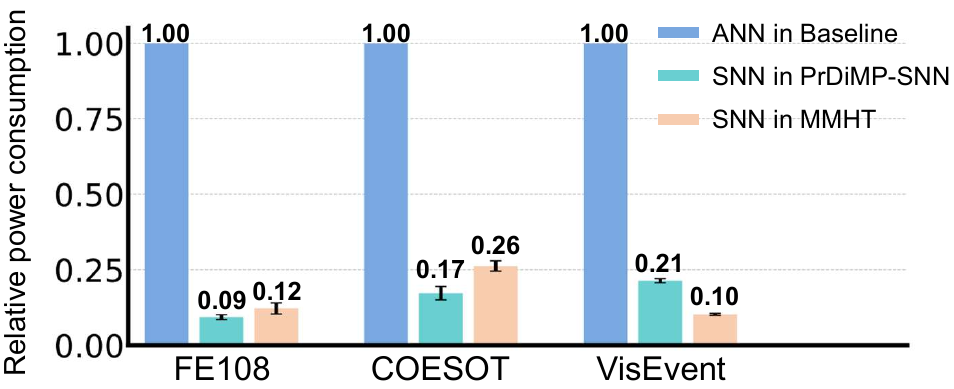}
\caption{The relative power consumption levels of the SNN backbones for a single modality and the fusion modality. }
\label{Figure6}
\end{figure}

\subsubsection{Analysis of the SNN Backbones for the Event Modality}
To demonstrate the influence of the SNN backbone in our MMHT model, we conduct experiments focusing on both performance and power consumption.

\begin{table*}[t]\footnotesize
\caption{Performance analysis of various fusion method.}
\centering
\begin{tabular}{c|p{2.0cm}<{\centering}p{2.0cm}<{\centering}|p{2.0cm}<{\centering}p{2.0cm}<{\centering}|p{2.0cm}<{\centering}p{2.0cm}<{\centering}}
	\hline
	\multirow{2}{*}{Method} &                   \multicolumn{2}{c|}{FE108}                    &                   \multicolumn{2}{c|}{COESOT}                   &                  \multicolumn{2}{c}{VisEvent}                   \\ \cline{2-7}
	& PR[\rm$\%$]                    & SR[\rm$\%$]                    & PR[\rm$\%$]                    & SR[\rm$\%$]                    & PR[\rm$\%$]                    & SR[\rm$\%$]                    \\ \hline
	Concat          & 89.14$\pm$.85                  & 61.70$\pm$.68                  & 71.26$\pm$.00                  & 65.57$\pm$.04                  & 69.05$\pm$.30                  & 54.60$\pm$.16                  \\
	Add           & 90.67$\pm$.06                  & 62.66$\pm$.03                  & 70.19$\pm$.12                  & 65.30$\pm$.07                  & 69.30$\pm$.24                  & 55.09$\pm$.25                  \\
	1$\times$1Conv  & 91.26$\pm$.29 & 62.34$\pm$.12 & 72.08$\pm$.21 & 65.63$\pm$.14 & 70.06$\pm$.19 & 54.91$\pm$.06 \\
	SE   & 89.58$\pm$.24 & 61.76$\pm$.17 & 69.76$\pm$.19 & 65.04$\pm$.16 & 68.92$\pm$.14 & 54.35$\pm$.12 \\
	TFF           & \textbf{93.62$\pm$.33}         & \textbf{62.97$\pm$.11}         & \textbf{74.03$\pm$.20}         & \textbf{65.81$\pm$.12}         & \textbf{73.26$\pm$.25}         & \textbf{55.10$\pm$.15}         \\ \hline
\end{tabular}
\label{table6}
\end{table*}

\begin{table*}\footnotesize
\centering
\caption{The effectiveness of different embedding dimension.}
\label{table7}
\begin{tabular}{l|p{2.0cm}<{\centering}p{2.0cm}<{\centering}|p{2.0cm}<{\centering}p{2.0cm}<{\centering}|p{2.0cm}<{\centering}p{2.0cm}<{\centering}}
	\hline
	\multirow{2}{*}{MFE} & \multicolumn{2}{c|}{$D_{{\rm dim}}=256$} &    \multicolumn{2}{c|}{$D_{{\rm dim}}=512$}     & \multicolumn{2}{c}{$D_{{\rm dim}}=1024$} \\ \cline{2-7}
	& PR[\rm$\%$]   & SR[\rm$\%$]              & PR[\rm$\%$]            & SR[\rm$\%$]            & PR[\rm$\%$]   & SR[\rm$\%$]              \\ \hline
	FE108                & 87.21$\pm$.78 & 57.94$\pm$.41            & \textbf{93.62$\pm$.33} & \textbf{62.97$\pm$.11} & 90.60$\pm$.67 & 60.44$\pm$.38            \\
	COESOT               & 61.14$\pm$.20 & 56.52$\pm$.09            & \textbf{74.03$\pm$.20} & \textbf{65.81$\pm$.12} & 70.11$\pm$.24 & 63.75$\pm$.42            \\
	VisEvent             & 66.99$\pm$.55 & 48.87$\pm$.28            & \textbf{73.26$\pm$.25} & \textbf{55.10$\pm$.15} & 72.07$\pm$.25 & 54.59$\pm$.11            \\ \hline
\end{tabular}
\end{table*}

\begin{table*}\footnotesize
\centering
\caption{The effectiveness of varying numbers of fusion iterations.}
\label{table8}
\begin{tabular}{l|p{2.0cm}<{\centering}p{2.0cm}<{\centering}|p{2.0cm}<{\centering}p{2.0cm}<{\centering}|p{2.0cm}<{\centering}p{2.0cm}<{\centering}}
	\hline
	\multirow{2}{*}{TMFF} &     \multicolumn{2}{c|}{\#Block=1}      &         \multicolumn{2}{c|}{\#Block=2}          & \multicolumn{2}{c}{\#Block=3} \\ \cline{2-7}
	& PR[\rm$\%$]    & SR[\rm$\%$]            & PR[\rm$\%$]            & SR[\rm$\%$]            & PR[\rm$\%$]   & SR[\rm$\%$]   \\ \hline
	FE108                 & 81.20$\pm$1.07 & 52.42$\pm$.60          & \textbf{93.62$\pm$.33} & \textbf{62.97$\pm$.11} & 82.53$\pm$.40 & 53.92$\pm$.14 \\
	COESOT                & 68.60$\pm$.13  & 62.60$\pm$.08          & \textbf{74.03$\pm$.20} & \textbf{65.81$\pm$.12} & 68.67$\pm$.32 & 61.02$\pm$.26 \\
	VisEvent              & 71.81$\pm$.25  & \textbf{55.86$\pm$.09} & \textbf{73.26$\pm$.25} & 55.10$\pm$.15          & 62.56$\pm$.27 & 44.31$\pm$.12 \\ \hline
\end{tabular}
\end{table*}

To demonstrate the superior performance of SNN in processing event data, we replace the SNN backbone of the event modality with a structurally identical ANN in both the PrDiMP-SNN and MMHT models. Specifically, the modified models are denoted as PrDiMP-ANN (for event-modal trackers) and MMHT-ANN (for fusion trackers), respectively. A detailed performance analysis is carried out, and the overall evaluation results, measured in terms of the PR and SR metrics, are presented in Tab.~\ref{table5}. Notably, across various datasets, the models utilizing SNN backbones demonstrate significantly superior performance to that of the models employing ANN backbones. Specifically, on the FE108 dataset, the SNN backbones contribute to $4.21\%$ and $4.81\%$ PR and SR improvement for the single-modal-based tracker, and $25.37\%$ and $18.98\%$ improvements for the multimodal-based tracker. On the COESOT dataset, the PR and SR improvement are $1.97\%$, $0.2\%$, $3.08\%$, and $3.26\%$, respectively. On the VisEvent dataset, the PR and SR improvements are $17.22\%$, $12.58\%$, $8.02\%$, and $4.15\%$, respectively.

Furthermore, prior studies have underscored the reduced energy consumption exhibited by SNNs~\citep{yao2023sparser, qu2023spiking, pei2019towards}. When compared with ANNs employing Multiply-Accumulate~(MAC) operations as their predominant computational units, SNNs utilizing small numbers of MAC operations solely in the input layers and mainly utilizing sparse Accumulate (AC) operations lead to notable power consumption reductions. A comprehensive quantitative analysis of the operations used by both the ANN and SNN backbones is presented in Tab.~\ref{table4}. Additionally, we provide the empirical power consumption values for both the MAC~($E_{\rm{MAC}}$) and AC~($E_{\rm{AC}}$) operations executed on the chip. Subsequently, we derive the theoretical energy consumption levels of ANN ($\Phi_{\rm{ANN}} $) and SNN ($\Phi_{\rm{SNN}} $) as follows:
\begin{equation}
\Phi_{\rm{ANN}} = E_{\rm{MAC}}\cdot MAC_{\rm{ANN}},
\label{eq17}
\end{equation}
\begin{equation}
\Phi_{\rm{SNN}} = E_{\rm{MAC}}\cdot MAC_{\rm{SNN}} + E_{\rm{AC}}\cdot AC_{\rm{SNN}} .
\label{eq18}
\end{equation}
Consequently, the relative energy consumption of the SNN backbones in relation to that of the ANN backbones can be defined as:
\begin{equation}
\eta =\frac{\Phi_{\rm{SNN}}}{\Phi_{\rm{ANN}}} .
\label{eq19}
\end{equation}
Obviously, a smaller value of $\eta$ means a lower energy consumption. 

In experiments, we randomly select five samples from various datasets to test the firing rates~(FRs) and statistic the corresponding MAC and AC operations. An illustration of the power consumption of the SNN backbones relative to that of identical ANN backbones~(denoted as ``ANN in Baseline'') is presented in Fig.~\ref{Figure6}. Due to the utilization of AC operations and sparse firing rates, the SNN backbones demonstrate a substantial reduction in power consumption. Specifically, on the FE108 and VisEvent datasets, the SNN backbones exhibit up to $0.90$ power savings. On the COESOT dataset, the ratios hover around $0.80$. Additionally, we also conduct a comprehensive examination to assess the energy consumption of the overall backbone within MMHT tracker. In contrast to utilizing Resnet+ANN as the backbone, the Resnet+SNN backbone demonstrates a notable energy conservation of approximately $0.10$~(with specific reductions of $0.12$ on FE108, $0.07$ on COESOT, and $0.12$ on VisEvent).

\subsubsection{Effectiveness of the TFF Fusion Method}
To validate the effectiveness of our proposed TFF feature fusion method, we conduct experiments encompassing an evaluation of several feature fusion approaches for comparative analysis purposes. Specifically, the following techniques are considered:~(1) concatenation~(referred to as 'Concat'), which involves the concatenation of the feature maps generated from the ANN and SNN backbones along the channel dimension; ~(2) addition~(referred to as 'Add'), which entails the fusion of feature maps through element-wise summation at the corresponding positions;~(3) pointwise convolution~(denoted as '1$\times$1Conv'), which is employed to consolidate features from diverse backbones by means of a 1$\times$1 convolution on individual pixels~\citep{howard2017mobilenets}; and~(4) squeeze-and-excitation attention block~(denoted as 'SE'), which is introduced to adaptively recalibrate the significance of each channel across feature maps from both modalities~\citep{hu2018squeeze}. The detailed results obtained based on diverse fusion methods are presented in Tab.~\ref{table6}. In the comparative assessment of the 'Concat', 'Add', '1$\times$1Conv' and 'SE' methods, '1x1Conv' exhibits superior performance across the majority of metrics, including PR on FE108, PR and SR on COESOT, and PR on VisEvent. Conversely, 'SE' yields suboptimal results, except for SR on FE108. These findings align consistently with those reported in a previous study~\citep{wang2021visevent}. However, the TFF method consistently exhibits superior performance across all metrics. Furthermore, in comparison with the single-modal trackers presented in Tab.~\ref{table3}, the multimodal trackers employing relatively simple feature fusion methods~('Concat', 'Add', '1$\times$1Conv' and 'SE') still demonstrate enhanced tracking performance. This outcome serves to underscore our earlier conclusion that multimodal data provide opportunities for achieving reliable object tracking.

Moreover, we conduct an in-depth analysis of two key components within the TFF module. (1) The embedding dimensionality $D_{{\rm dim}}$ in the Multimodal Feature Emdedding~(MFE) plays a crucial role in influencing the sparsity of the features within the transformation space. Accordingly, by varying the size of $D_{{\rm dim}}$, we demonstrate its impact on the TFF fusion method, as presented in Tab.~\ref{table7}. The experimental findings reveal that deploying a smaller embedding dimensionality leads to a notable performance decline, which is particularly evident on the COESOT dataset, with a reduction of approximately $10\%$. On the other hand, utilizing a larger value still results in performance degradation; however, the tracker maintains a relatively commendable performance level. We speculate that both excessively small and excessively large feature sparsity values can yield unfavorable outcomes for the model. Specifically, an excessively small $D_{{\rm fc}}$ may result in information loss during the feature transformation process, while excessive feature redundancy may impose a burden on the effectiveness of model training. (2) We also discuss the number of Transformer-based Multimodal Feature Fusion~(TMFF) modules to elucidate the correlation between tracker performance and the iterative fusion process. The outcomes of the experiments are presented in Tab.~\ref{table8}, revealing the noteworthy influences of varying numbers of fusion iterations on the ultimate performance of the model. For the FE108 and COESOT datasets, the model attains its optimal performance with 2 fusion iterations. However, in the case of the VisEvent dataset, the performance of the model demonstrates approximate equivalence between 1 and 2 fusion iterations. Combining the experimental findings observed in this section, the MMHT model proposed in this manuscript is characterized with the following parameters $D_{{\rm dim}}$=512 and \#Block=2. 

\subsection{Multimodal Feature Fusion on the MMHT}
\begin{figure}[t]
	\centering
	\includegraphics[width=1.0\columnwidth]{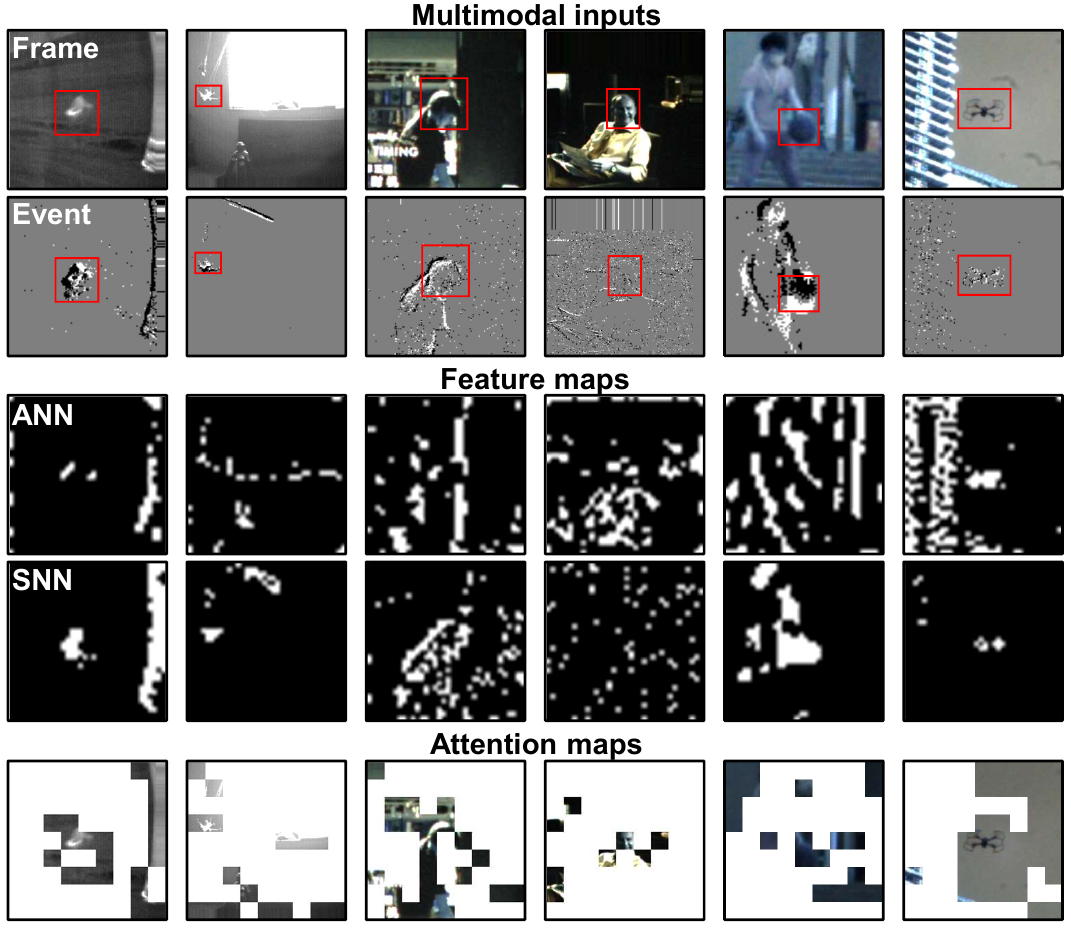} 
	\caption{Visualization of the produced feature maps and attention maps. The feature maps are randomly selected from representative channels in the high-level feature $F_{h}^{i}$ and $G_{h}^{i}$ while the attention maps are transformed into masks and overlaid on the frame inputs to facilitate a visual interpretation. The red boxes indicate the ground truths.}
	\label{Figure7}
\end{figure}

To demonstrate the operational mechanisms of the proposed MMHT model, we visualize the feature maps extracted from the hybrid backbone and the attention maps generated by the fusion module. Specifically, to compare the functional disparities of backbones in capturing visual cues, we randomly select feature maps from the representative channels in $F_{h}^{i}$ and $G_{h}^{i}$, respectively. Regarding the fusion module, we initially compute the average of the fusion embeddings $G_{\rm h,embed}^{i}$ and $F_{\rm h,embed}^{i}$ along the dimensional direction. Subsequently, attention maps are derived by retaining only the significant regions where the value exceeds $0.5$. As shown in Fig.~\ref{Figure7}, the ANN backbone effectively captures intricate texture features from the frame modality, while the SNN backbone exhibits heightened sensitivity toward moving objects. By integrating multimodal features, the fusion module precisely focuses its attention on the target object and crucial background information, thereby enhancing the reliability of object tracking. To a certain extent, our observation demonstrates that the MMHT model can effectively fuse multimodal features across different domains.


\section{Conclusion}
In this paper, we proposed a novel MMHT model, aiming to exploit the potential of diverse visual modalities to achieve reliable single object tracking. Specifically, we designed a hybrid backbone that enables the extraction of features from multiple visual modalities. By aligning and mapping visual features from different modalities into a unified visual feature space, we employed an enhanced transformer-based module to effectively fuse the  discriminative features across different domains. The performance of our proposed approach was evaluated on benchmark datasets, and the results demonstrated the superiority of the MMHT model over other state-of-the-art models. 

In subsequent ablation experiments, we further demonstrated that the proposed fusion model can effectively integrate crucial visual cues from different visual modalities, achieving reliable object tracking across various challenging attributes. The backbones were analyzed in terms of both their effectiveness and energy consumption, thus explaining why it is necessary to use SNNs for extracting features from event modality inputs. Furthermore, in contrast with various multimodal fusion strategies, our MMHT model consistently upheld its superiority.

In future work, we will continue to refine our MMHT model to enhance its tracking performance in more intricate scenarios and improve its tracking speed. Furthermore, we will direct our attention toward addressing challenging multi-object tracking tasks.

\section*{Acknowledgments} 
This work was supported in part by the STI 2030 Major Project under grant 2022ZD0208500, in part by the National Key Research and Development Program of China under grant 2023YFF1204202, in part by the National Natural Science Foundation of China under grant  82072011 and in part by Sichuan Science and Technology Program under grant 2024NSFJQ0004.

\bibliographystyle{elsarticle-harv} 
\bibliography{ref}

\end{document}